\documentclass[conference]{IEEEtran}
\usepackage{stmaryrd}
\usepackage{amsfonts}

\usepackage{graphicx,times,amsmath,amssymb,subfigure} 
\usepackage{flushend}
\IEEEoverridecommandlockouts    

\begin{document}

\title{\ \\ \LARGE\bf A Subpixel Registration Algorithm for Low PSNR Images \thanks{This work was supported by National Natural Science Foundation of China (11163004,11003041), and Open Research Program of Key Laboratory of Solar Activity of Chinese Academy of Sciences (KLSA201205, KLSA201221).}\thanks{
S Feng is with  Computer Technology Application Key Lab of Yunnan Province, Kunming University of Science and Technology, Chenggong Kunming, China, 650500, and Key Laboratory of Solar Activity, National Astronomical Observatories, Chinese Academy of Sciences, Beijing 100012(email:ynkmfs@gmail.com). }\thanks{ L Deng is with National Astronomical Observatories/Yunnan Astronomical Observatory, Chinese Academy of Sciences, Kunming, China 650011 and Graduate University of Chinese Academy of Sciences,Beijing, China 100049.}\thanks{ G Shu is with  Computer Technology Application Key Lab of Yunnan Province, Kunming University of Science and Technology, Chenggong Kunming, China, 650500, and Key Laboratory of Solar Activity, National Astronomical Observatories, Chinese Academy of Sciences, Beijing 100012.}\thanks{F Wang and H Deng are with Computer Technology Application Key Lab of Yunnan Province, Kunming University of Science and Technology, Chenggong Kunming, China, 650500.}\thanks{K Ji, corresponding author, is with Computer Technology Application Key Lab of Yunnan Province, Kunming University of Science and Technology, Chenggong Kunming, China, 650500.(phone: +86-8713340576; fax: +86-8713340576; email: jkf@cnlab.net) }}

\author{Song Feng, Linhua Deng, Guofeng Shu, Feng Wang, Hui Deng and Kaifan Ji}


\maketitle

\begin{abstract}
This paper presents a fast algorithm for obtaining high-accuracy subpixel translation of low PSNR images. Instead of locating the maximum point on the upsampled images or fitting the peak of correlation surface, the proposed algorithm is based on the measurement of centroid on the cross correlation surface by Modified Moment method. Synthetic images, real solar images and standard testing images with white Gaussian noise added were tested, and the results show that the accuracies of our algorithm are comparable with other subpixel registration techniques and the processing speed is higher. The drawback is also discussed at the end of this paper.
\end{abstract}


\section{Introduction}

\PARstart{M}{any} image processing applications require registering multiple overlaying the same scene images taken at different times, such as image fusion and super-resolution in astronomical observation, remote sensing, biomedical imaging and so on.  Most astronomical objects are extremely faint and very low contrast and emit light in certain specific wavelengths. These characteristics create many problems in image processing, especially in images registration for increasing spatial solution or signal-to-noise ratio by accumulating multiple frames. For example, hundreds of weak solar Magnetic Field images are needed to be registered for decreasing the effects of atmospheric turbulence, wind and the tracking accuracy of the telescope itself in real-time. Due to the short time interval of images, the peak signal-to-noise ratio (PSNR) is less than 20dB usually.

A wide variety of algorithms can be found for solving image registration problem. The techniques can be classified in feature-based and area-based (or intensity-based)\cite{cit:1}. Feature-based techniques aim to detect the pairwise spatial relation or various descriptors of features, which represented by the control points (CPs), such as points themselves, end points or centers of line, centers of gravity of regions and so on. This approach can handle complex between-image distortions, but requires that the register images must have many prominent details. However, the edges of solar image texture are fuzzy and smooth. CPs for feature-based techniques are hard to find.

The basic idea of area-based matching techniques is to achieve minimizing the compared images' squared error, maximizing their correlation without any structural analysis. Typical examples include methods using image correlation functions, which can be calculated in spatial or frequency domain. Over the years, various high-accuracy image registration algorithms based on correlation techniques have been developed, especially in processing subpixel image registration.

Usually, subpixel registration covers estimate of subpixel translation, rotation and scaling\cite{cit:2}. In this paper, we are primarily concerned with evaluation of 2D rigid translation and propose a fast subpixel shift registration algorithm based on image correlation for low PSNR images.

Cross-correlation (CC) and phase-correlation (PC) algorithms are two well-known images registration techniques in frequency domain\cite{cit:1},\cite{cit:3}. Both of them take Fourier Transform (FT) first and calculate the cross-power or phase of the cross-power spectrum, then take Inverse Fourier Transform (IFT) to generate correlation surface. The coordinate of the maximum on the surface can be used as an estimate of the horizontal and vertical components of translation between two images. Comparing with CC, PC uses the normalized cross-power spectrum instead of cross-power spectrum, so it provides a distinct sharp peak at the point of registration whereas CC yields several broad peaks\cite{cit:1}. If pixel level registration is satisfactory for applications, locating the PC peak is equivalent to finding the displacement at pixel level.

In order to obtain subpixel accuracy and keep using the same algorithms of locating the peak coordinate of PC, traditional approach is to compute an upsampled correlation between the shifted image and reference image. This approach not only significantly increases the size of IFT that need be computed as the required accuracy increases (for example, registration within 1/100 of a pixel requires computational cost 10000 times larger than original images being correlated), but also strongly associates with interpolation techniques\cite{cit:4}. Literature\cite{cit:5} proposed a more efficient implementation using nonlinear optimization algorithm to reduce computational time and memory requirement. Literature\cite{cit:6} discussed different solutions to address the interpolation problem in image registration based on high degree B-spline. Since the signal power in the cross or phase correlation is not concentrated in a single peak, but rather in several coherent peaks mostly adjacent to each other, the approaches of estimating the peak position on subpixel level without upsampling images were developed during last ten years. In phase correlation, the peak is shape and yields a 2D Dirichlet kernel, which can be closely approximated by a $sinc$ function. Some different fitting or interpolation functions in a given neighborhood were proposed by \cite{cit:7}-\cite{cit:10}, such as fitting of triangle, 1-D or 2-D Gaussian or $sinc$ functions to achieve high accuracy subpixel registration. Base on the Fourier Shift Theorem, the shift parameters can be computed by slope of phase difference of two images in the Fourier domain. Several algorithms were proposed by \cite{cit:11}-\cite{cit:14} as well. And in the literature\cite{cit:15}, a method based on gradient correlation(GC) was presented and compared with PC. Literature\cite{cit:11},\cite{cit:16} evaluated the subpixel registration accuracies of different phase-based method.

\section{Proposed Algorithm}
For achieving high accuracy result, our method is based on Coarse-to-Fine approach\cite{cit:17}. In the first stage, a coarse pixel level displacement is obtained and image is shifted in pixel level. In the second stage, a finer approximation is performed to assess subpixel translation. Since pixel level registration is quite simple and easy implemented, the refinement of coarsely registered images to subpixel accuracy is our focus in this paper.

Although popular, phase correlation techniques have some disadvantages. It is more sensitive to noise than direct CC, both for low-pass and high-pass inputs. Sharpening the correlation peaks comes at the expense of increased sensibility to noise of the computed maximum position\cite{cit:3}. Barbara Zitova et al.\cite{cit:1} indicated that in a general case PC could fail in case of multimodal data as well. The peak of cross correlation function is rather broad, and there are more surrounding pixels could be involved to estimate the displacement instead of a few of pixels in PC surface. Therefore, cross correlation is selected as our basic approach.

In our application, since low computational complexity is critical, we try to avoid any fitting or interpolation, and calculate centroid of peak directly in our algorithm. In astronomical image processing, Modified Moment is a popular digital centering algorithm\cite{cit:18} and has been widely applied in measurement of star positions in CCD images. Comparing with classic Gaussian fit, it can provide same accuracy with very high speed.

In this method, a threshold level is set and only those count levels in the data array that are above this threshold are considered in the centering process. Specifically, once the threshold $b$ is determined in the $I(x,y)$ data array, each pixel in the array is thresholded according to the following precepts:

\begin{equation}
 \begin{array}{ll}
     I'(x,y)=I(x,y)-b & \textrm{if $I(x,y)> b$ }\\
     I'(x,y)=0 & \textrm{if $I(x,y)<b$ }
    \end{array}
 \label{eq:eq1}
\end{equation}

Then, the centroid of the peak is set equal to the first moment of the marginal distribution.
\begin{equation}
            x_{c}=\sum_{x} xI'(x,y)/\sum_{x} I'(x,y)  \\
\end{equation}
\begin{equation}
            y_{c}=\sum_{y} yI'(x,y)/\sum_{y} I'(x,y)  \\
\end{equation}

Single star peak looks like cross correlation surface, which is surrounded by a few of coherent peaks. We apply Modified Moment in the estimate of centroid of cross correlation surface. First, select a radius around the maximum of the peak, and set the threshold $b$ by the minimum value inside the circle. Second, calculate each pixel value above the threshold by Eq.(1). Third, determine the centroid with the first moment by Eq.(2) and Eq.(3). The centroid coordinates equals the displacement between two images.

Obviously, the algorithm is based on the assumption that the maximum of the peak is at same position as the centroid. It means the correlation value above the threshold should be symmetrical in the horizontal and vertical directions. The assumption is acceptable in the small center area around the cross correlation peak. In our experiment, the radius was set to 3 pixels.

\section{Experimental Setup}
In order to test the performance of the accuracy and robustness against noise of our algorithm, two sets of images with subpixel translation were generated without any interpolation technique.

In date set \uppercase\expandafter{\romannumeral1}, the reference image $I_{r}(x,y)$ and shifted images $I_{s}(x,y)$ were generated by analytical expressions, and they could be calculated by a liner combination of several two-dimensional Gaussian functions.
\begin{figure}[htp]
\centerline{\includegraphics[width=3.7in]{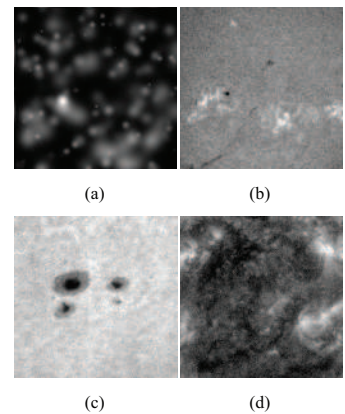}}
\caption{The images of data set I and II were used in algorithm performance testing.  (a) Synthetic image. (b) 656 nm wavelength of solar images observed on the earth. (c) 532 nm wavelength of solar images observed on the earth. (d) 17 nm wavelength of solar images observed on the space.}
\label{fig_1}
\end{figure}
{\begin{figure*}[thb]
        \begin{center}
        \subfigure[]{
        \includegraphics[width=6cm]{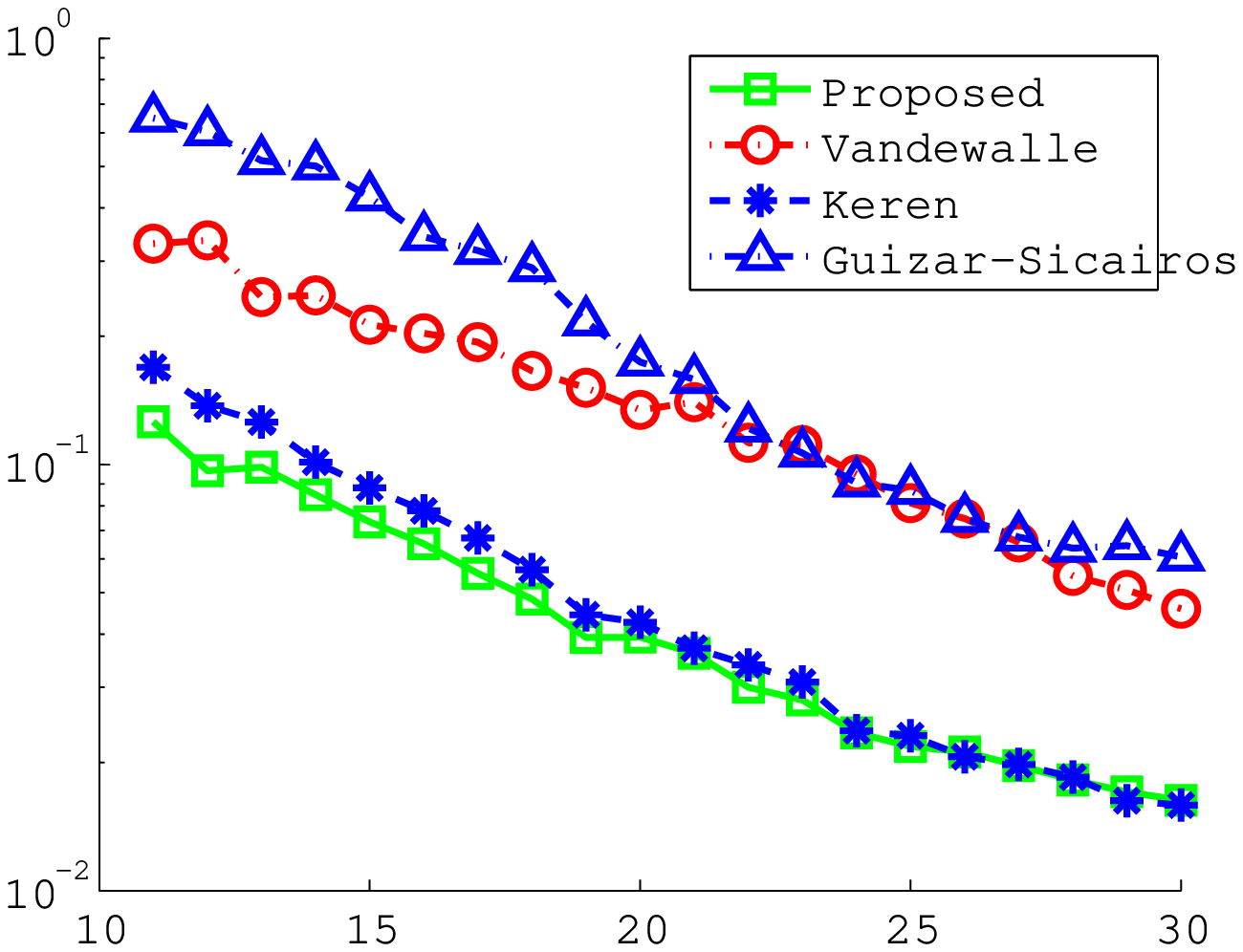}}
        \subfigure[]{
        \includegraphics[width=6cm]{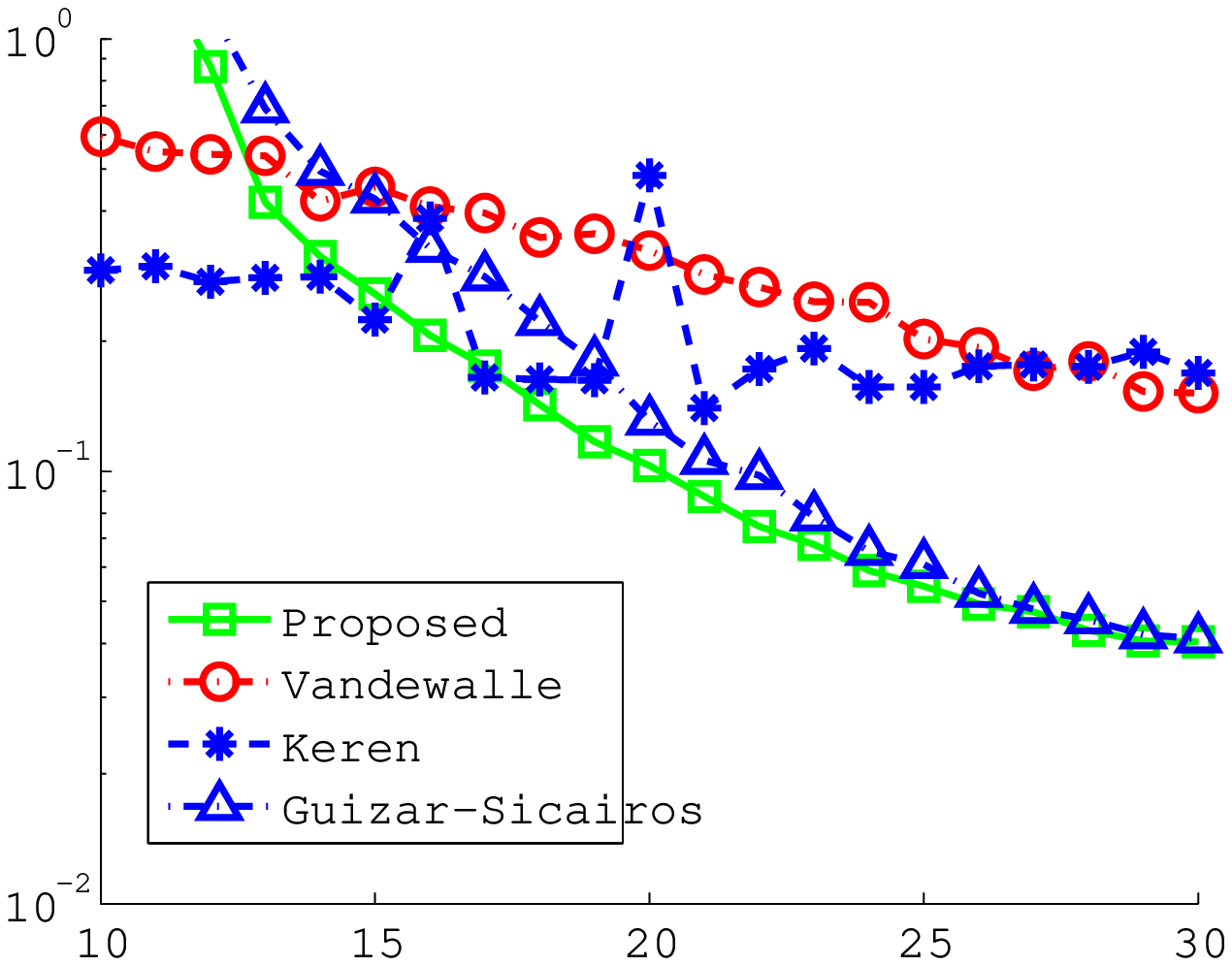}}
        \subfigure[]{
        \includegraphics[width=6cm]{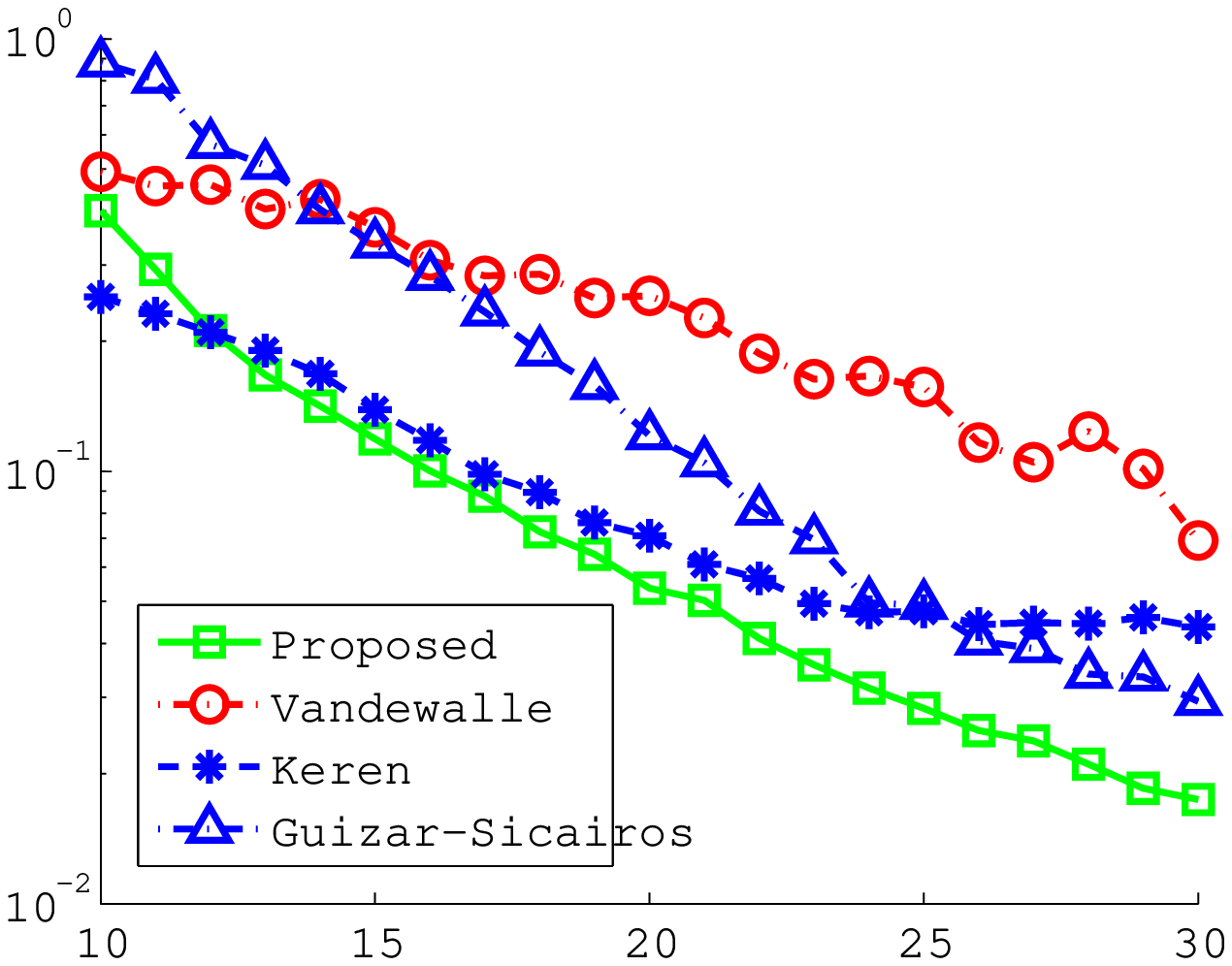}}
        \subfigure[]{
        \includegraphics[width=6cm]{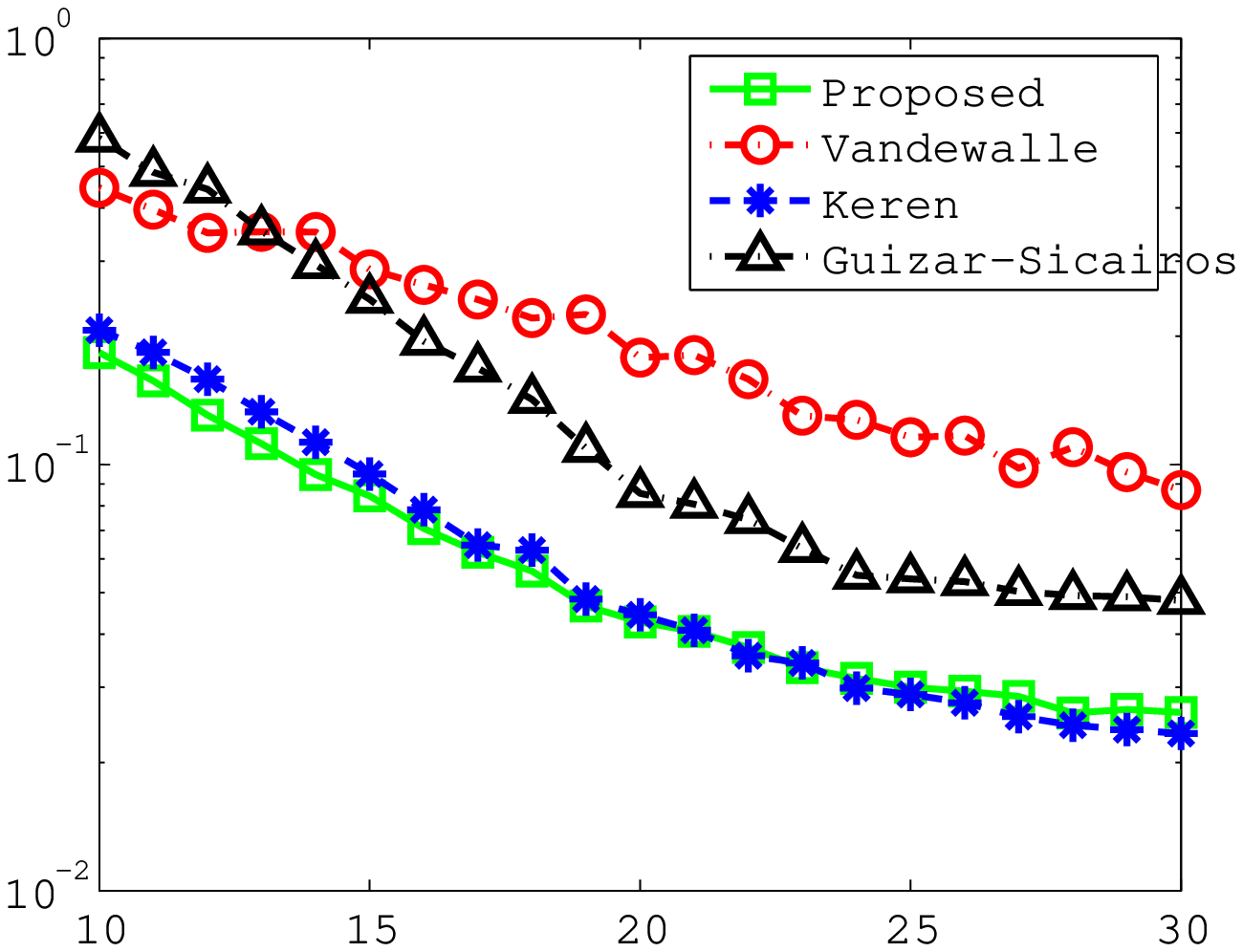}}
        \caption{Results of four algorithms on the test images which are shown on Fig.1, where X axis represents the PSNR in dB and Y axis represents the accuracy of registration on specific PSNR. (a) Synthetic image.  (b) 656 nm wavelength of solar images observed on the earth. (c) 532 nm wavelength of solar images observed on the earth. (d) 17 nm wavelength of solar images observed on the space.}
        \label{fig:qq}
        \end{center}
\end{figure*}}

\begin{equation}
    I_{r}(x,y)=\sum_{i}A_{i}exp\{-\frac{(x-\bar{x_{i}})^{2}+(y-\bar{y_{i}})^{2}}{2\sigma_{i}^{2}}\}   \\
\end{equation}
\begin{equation}
    I_{s}(x,y)=\sum_{i}A_{i}exp\{-\frac{(x-\bar{x_{i}}-dx)^{2}+(y-\bar{y_{i}}-dy)^{2}}{2\sigma_{i}^{2}}\} \\
\end{equation}

Here the coefficient is $A$ the amplitude, $\bar{x}$ and $\bar{y}$ are the center and $\sigma$ is the variance of each Gaussian function. Two hundred Gaussian functions were applied in our synthetic image, and their $\bar{x}$, $\bar{y}$ and $\sigma$ were generated by uniform distribution on the interval [0, 128] and [1, 6]. The image is shown on Fig.1(a). $dx$ and $dy$ define the applied transformation corresponding to the shift through the $x$ and $y$ directions, which are chosen from a uniform distribution on the interval [-0.5, 0.5]. The advantage of this method is that an analytical translation field can be straight forward applied to deform reference image\cite{cit:17}.  One hundred images of 128$\times$128 were generated with different subpixel translation.

{\begin{figure*}[thb]
        \begin{center}
        \subfigure[]{
        \includegraphics[width=6cm]{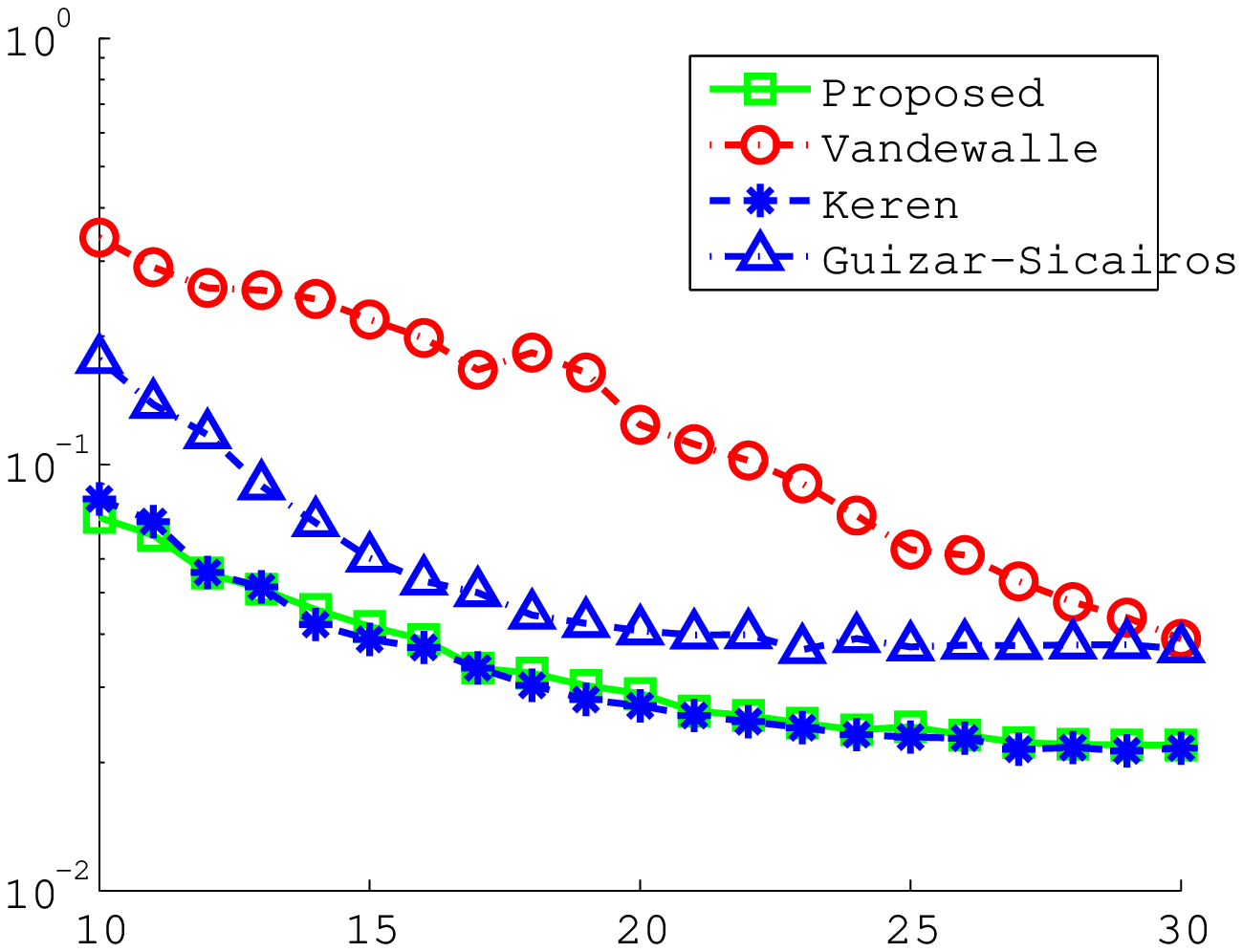}}
        \subfigure[]{
        \includegraphics[width=6cm]{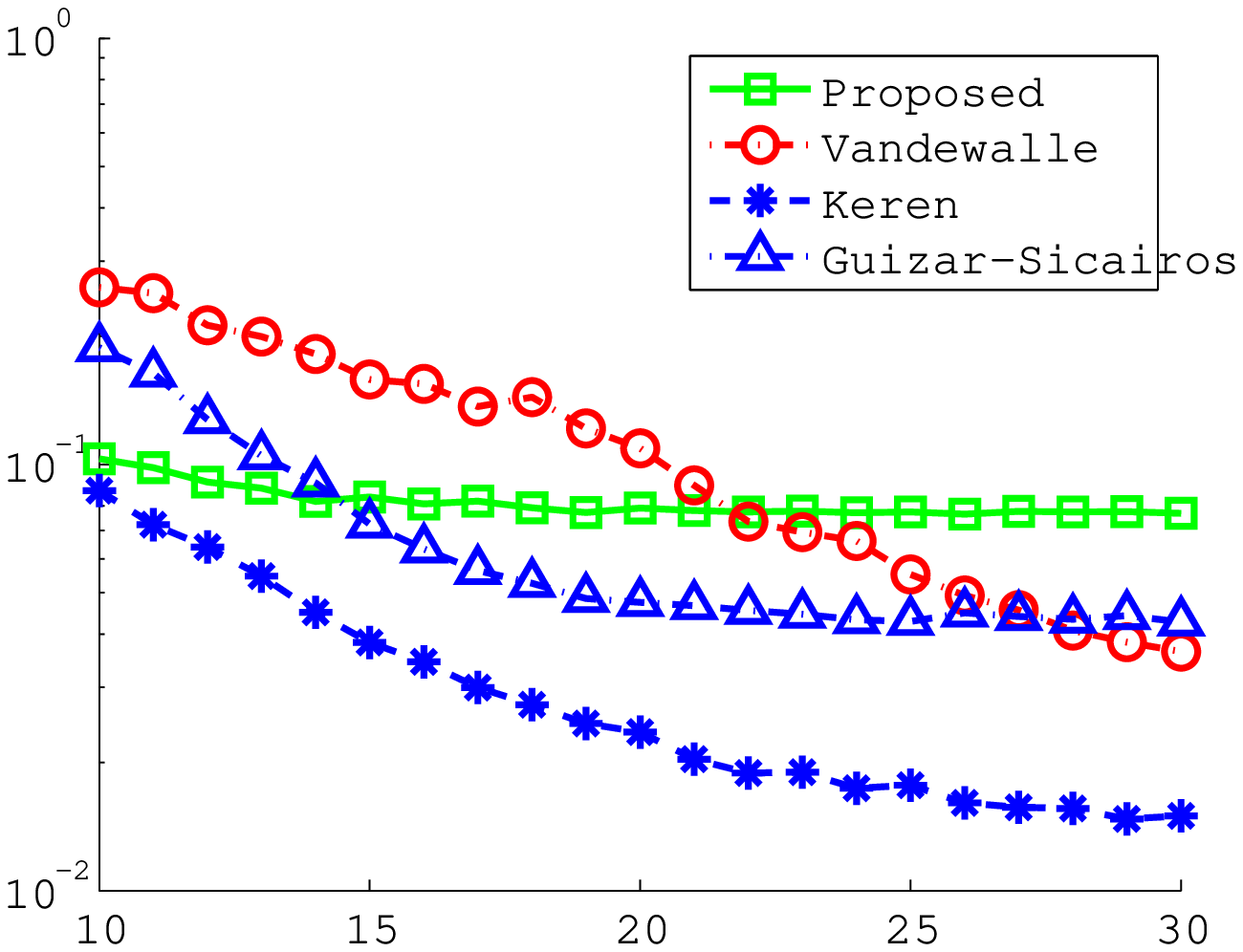}}
        \subfigure[]{
        \includegraphics[width=6cm]{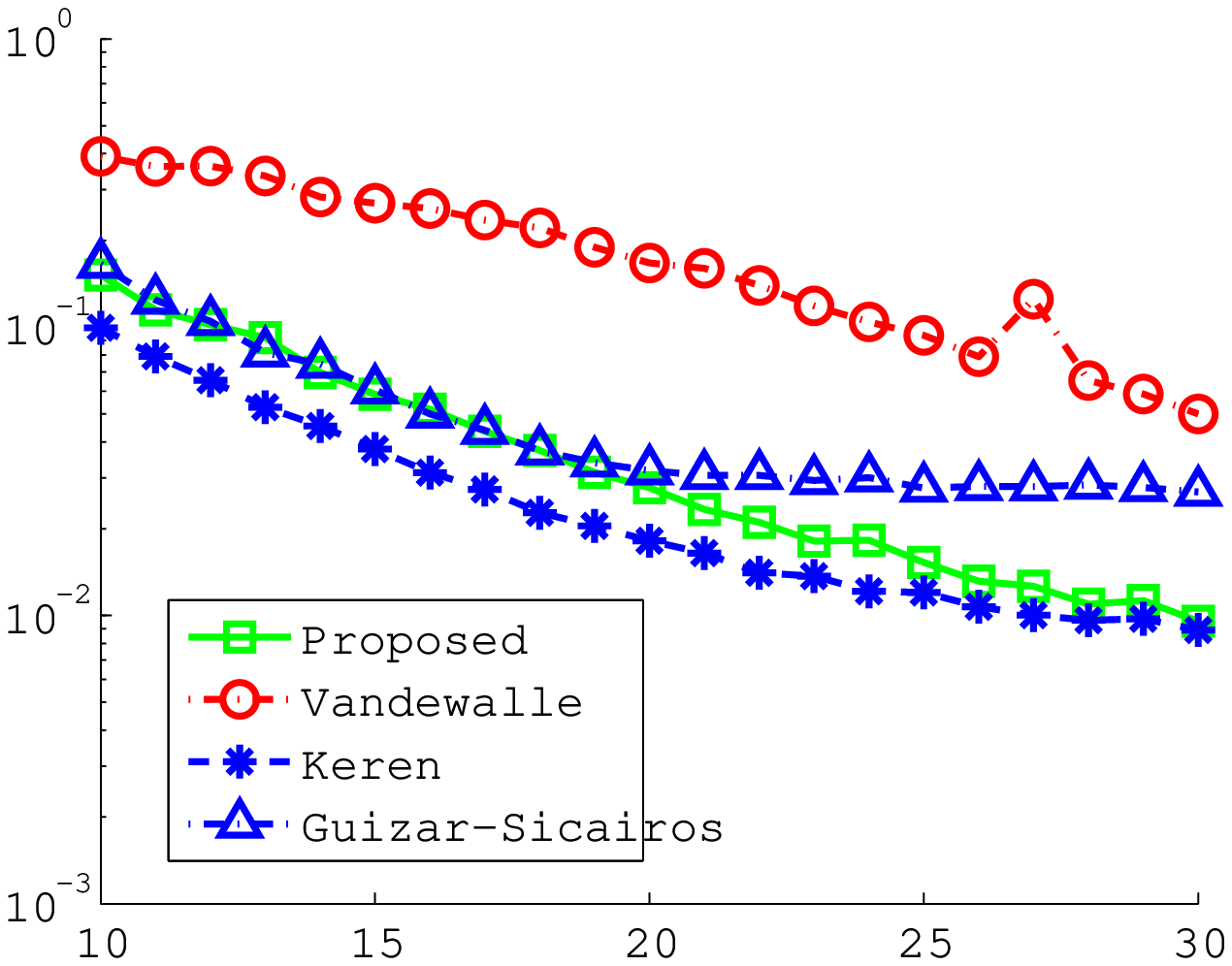}}
        \subfigure[]{
        \includegraphics[width=6cm]{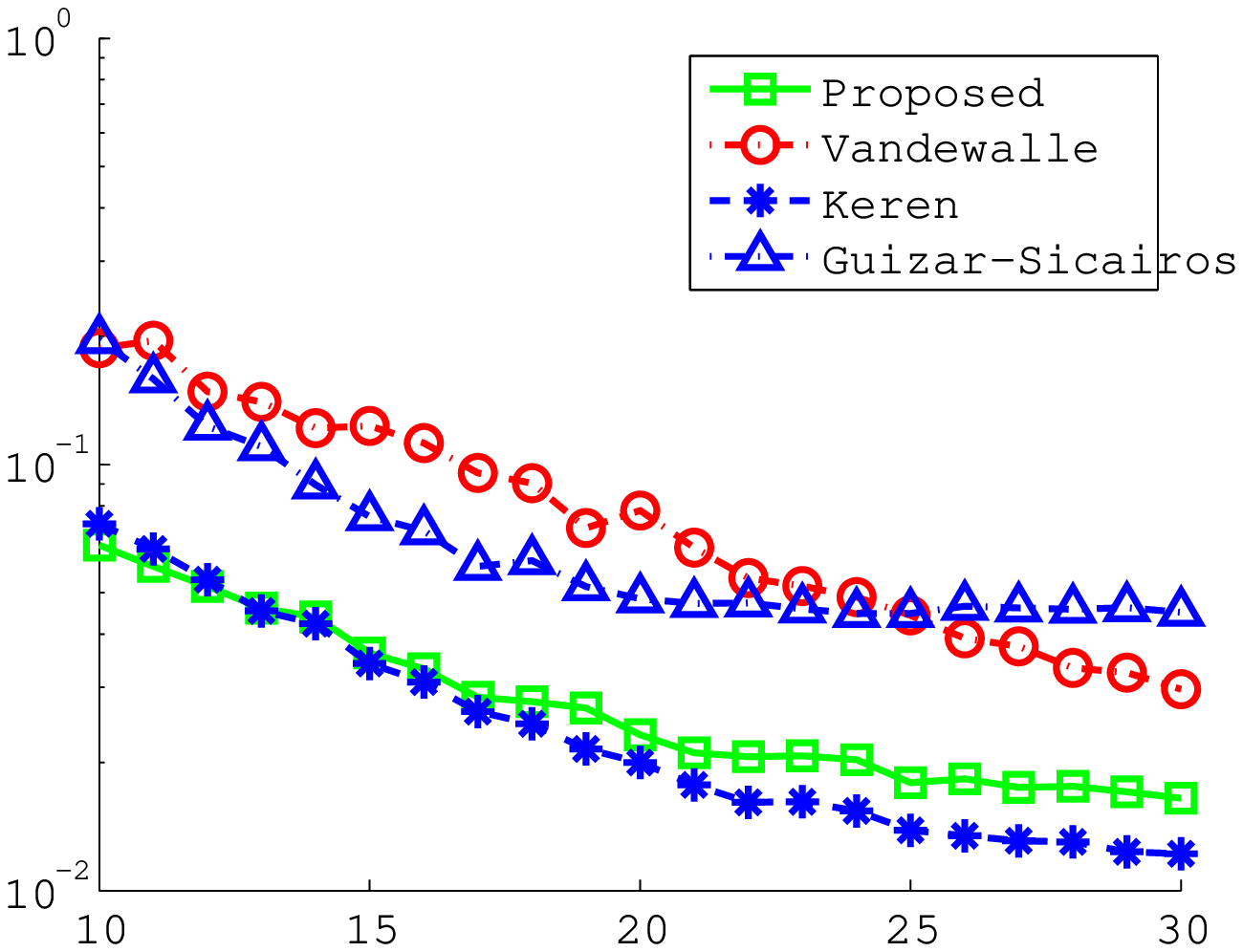}}
        \caption{Results of four algorithms on the standard testing images, where X axis represents the PSNR in dB and Y axis represents the accuracy of registration on specific PSNR. (a) Camera Man (b) Lena (c) Pentagon (d) Barbara}
        \label{fig:qq}
        \end{center}
\end{figure*}}

In data set \uppercase\expandafter{\romannumeral2}, we followed the scheme described in \cite{cit:14},\cite{cit:16}, the reference and register images were generated by filtering and downsampling a single high-resolution image 2200$\times$2200. The downsampled images were shifted with respect to each other by integer amounts and in the high-resolution grid. After downsampling by a factor D of in each dimension, the relative shifts became fractional shifts of size. After selecting a reference image, all the others were a complete set of translated versions of the reference image within [-0.5, 0.5] pixel in both direction, with exact subpixel displacements in portions of 1/D. We chose D=16, and generated 289 images of 128$\times$128. Three real typical solar images shown on Fig.1 (b)-(d) were employed as testing images, which were observed on different times, locations and wavelengths.

Then the white Gaussian noises with different variances were added to both reference and shifted images, and the variances of PSNR were from 10dB to 40dB on the interval of 2dB.
\section{Results and Discussion}
For testing the performance, our proposed method applied to data set I and II, and three different subpixel registration algorithms that were proposed by M. Guizar-Sicairos et al.\cite{cit:5} based on upsampled cross correlation, by P.Vandewalle et al.\cite{cit:11} based on the slope of the phase difference in frequency Domain, and by D.Keren et al.\cite{cit:19} based on Taylor expansions in spatial domain were applied as well.

Fig.2 shows the accuracies of four algorithms on the test images versus PSNR, where X axis represents the PSNR in dB and Y axis represents the accuracy of registration on specific PSNR. Since the real subpixel level translations of both data sets were known exactly, the accuracies of different algorithms could be estimated by standard deviation of their results. The conclusion is that our proposed algorithm provides excellent performance both for data set I and II, and the approach by D.Keren et al. also shows very good results except image Fig.1(b).

Besides data set I and II, standard testing images were processed too, include Barbara, Camera Man, Lena and Pentagon. And the results are shown on Fig.3.

Except image Lena, the proposed algorithm provides acceptable subpixel accuracy compared with other algorithm especially in low PSNR. However, the results also show that the performance depends on the autocorrelation of images. If the cross correlation surface is too flat, it is hard to get high accuracy of subpixel registration. Fortunately, in our application, autocorrelation of the images is low.

The calculation speed of our proposed algorithm is higher than others. Under our experiment environment (Lenovo X200s notebook with 1.86GHz Intel Core2 Duo CPU, 4G memory, Windows 7 and Matlab 7.0), the measurement of the subpixel translation between two 128$\times$128 images took 5 milliseconds by our code, and 66 milliseconds by M.Guizar-Sicairos et al., 8 milliseconds by P.Vandewalle et al., and 152 milliseconds by D.Keren et al. This is important for us, because we would like to get real-time image registration.

\section{Conclusions}
The methodologies we describe in this paper consist of three steps. First, calculate the cross correlation surface. Second, set the threshold by the minimum value inside a small circle around the maximum of the peak. Third, estimate the centroid above the threshold by the Modified Moment method. Based on the results of processing both synthetic image and real images, our algorithm presents comparable high accuracy for noisy image subpixel registration. Neither upsample nor interpolation techniques are involved, so the speed of the proposed algorithm is very high and could be implemented in real-time image processing. The drawback of our algorithm is that we might not get the prefect result if the autocorrelation of image is high, since cross correlation technique is applied. Obviously, the performance depends not only on PSNR, but also on the construction of images. It is hard to develop an algorithm which could provide perfect accuracy for all images. Considering of our application, the proposed algorithm is a better solution than others.

\section{Acknowledgments}
We appreciate the support from National Natural Science Foundation of China (11163004, 11003041), and Open Research Program of Key Laboratory of Solar Activity of Chinese Academy of Sciences (KLSA201205, KLSA201221).



%

\def\V{\rm vol.~}
\def\N{no.~}
\def\pp{pp.~}
\def\Pot{\it Proc. }
\def\IJCNN{\it International Joint Conference on Neural Networks\rm }
\def\ACC{\it American Control Conference\rm }
\def\SMC{\it IEEE Trans. Systems\rm , \it Man\rm , and \it Cybernetics\rm }

\def\handb{ \it Handbook of Intelligent Control: Neural\rm , \it
    Fuzzy\rm , \it and Adaptive Approaches \rm }

\end{document}